# MIXED-LEVEL KNOWLEDGE REPRESENTATIONS AND VARIABLE-DEPTH INFERENCE IN NATURAL LANGUAGE PROCESSING

MICHAEL HESS

*University of Zurich, Dept. of Computer Sciences*
*CH-8057 Zurich, Switzerland*
`hess@ifi.unizh.ch`

ABSTRACT

A system is described that uses a mixed-level knowledge representation based on standard Horn Clause Logic to represent (part of) the meaning of natural language documents. A variable-depth search strategy is outlined that distinguishes between the different levels of abstraction in the knowledge representation to locate specific passages in the documents. A detailed description of the linguistic aspects of the system is given. Mixed-level representations as well as variable-depth search strategies are applicable in fields outside that of NLP.

## 1. Outline

Most knowledge representation schemes used in AI (and, in particular, in Natural Language Processing) are homogeneous. One application that makes obvious the limitations of a homogeneous representation space is fact retrieval from natural language texts. Our understanding of the semantics of natural language is still so incomplete that the representation of the content of natural language texts must either remain fragmentary, or allow for expressions of varying degrees of abstractness to occur in the same representation scheme. We present, in the context of passage retrieval, a mixed-level representation scheme based on standard Horn Clause Logic and outline a variable-depth evaluation strategy to be used over this type of representation. The combination of such a multi-level representation and a search strategy that is sensitive to the resulting variability in granularity is of general interest for the design of knowledge representation schemes.

## 2. The Problem

The amount of textual information available today in machine-readable form makes it increasingly difficult to locate relevant information reliably and efficiently. With the number of machine readable documents accessible in contemporary data bases or over the networks going into the millions, even the best search systems based on traditional Information Retrieval (IR) methods overwhelm the user, in many cases, with thousands of documents. It would be extremely useful to have text based *fact retrieval systems* or, even more ambitiously, text based *question answering systems*. However, modeling the deep understanding of unrestricted text needed for these applications is still beyond our technical capabilities. This is why we are developing, as an intermediate solution, a system that is capable of taking us to the exact place in a natural language text that is relevant to a query (i.e. a passage retrieval system).

Work on this paradigm has just begun, mainly by researchers in the field of IR ([1]). However, it is doubtful that IR methods will be very useful in this new context due to their lack of precision. One of the reasons for this lack lies in the fact that these systems ignore almost all the linguistically relevant information in documents beyond the bare lexical skeleton. Thus *both* of the questions

1) **?- Logic for natural language analysis**
2) **?- Languages for the analysis of natural logic**

would be converted, under a keyword-based approach, into a query

```
?- logic & natural & language & analysis
```

and hence return the same documents although the original queries are far from synonymous. By ignoring function words, morphology, and word order, the keyword extraction procedure loses all the syntactic structure in the natural language input, and consequently all the semantic information that is encoded in natural language through syntax becomes unavailable. Any amount of statistical sophistication applied to the bare keywords will not be able to recover the information that was thrown away when extracting the keywords. This effect makes itself felt particularly clearly when one tries to use standard IR techniques to the goal of passage retrieval.

The goal of this paper is to show how a combination of well known techniques from Natural Language Processing (NLP) and AI, together with the less widespread techniques of mixed-level knowledge representation and variable-depth evaluation strategies, can overcome, in part, the performance related problems that would otherwise mar a fully-fledged NLP based passage retrieval system.

## 3. The use of linguistic information in Information Retrieval so far

Most of the IR systems described in the literature that make use of linguistic information at all fall into the category of syntax driven automatic indexing. In such systems, syntax structures are merely used to derive phrase descriptors, i.e. multi-term descriptors which are then used in the standard way. However, retrieval results are not significantly better than with standard approaches ([2]). Some systems now use the syntactic structures *themselves* as descriptors. In such systems, the syntax structure of a query will have to match the syntax structure of some expressions in a candidate documents *directly*, i.e. without the intermediary step of phrase descriptors (e.g. [3], [4]).

However, if we use full syntax structures as descriptors, we run into the problem that there will almost never be an exact match between the syntax structure of query and (parts of the) documents. First, the same state of affairs can almost always be described in a variety of ways and, in particular, widely different syntactic structures can be used to express the same meaning. For a direct syntax matching approach this has the consequence that we can not even hope to find an *immediate* match between the syntax structure of the query and the syntax structure of (part of) a candidate document. Second, queries normally denote *supersets* of what the relevant documents denote. If our query is ''programming languages'' we want to retrieve (among others) all the documents about ''*object-oriented* programming languages''. Again, the syntax structures of the query and of the expressions making up the documents will be different, and we need some kind of syntactic correlate of the semantic superset/subset relation.

## 4. Using logic as search and index language

An alternative solution to the problems outlined above that was implemented in a prototype passage retrieval system, LogDoc, starts from the idea that the problem underlying syntactic variability is really a mismatch between a syntactic structure and a semantic expression: Although their *syntactic realisations* are different, the *meanings* of two phrases may be the same, and it is *this* relationship that we are ultimately interested in. The basic idea of LogDoc is, thus, to use logic, with a simple ontology, as a knowledge representation language. We translate documents into logical axioms with back-pointers to the source text, add the axioms to an incrementally growing logical data base, translate queries into theorems, and by proving the latter over the former we are able to retrieve the original documents answering the query. - Key to the success of this procedure is the fact that we can express in logic certain complex relationships between word senses that we cannot represent with standard IR representation schemes. If we use First Order Logic as representation language a document fragment like 3

   3) **A structure sharing representation of language for unification based grammar formalisms**

might, for instance, become

```
3a) representation(R,L) ∧ language(L) ∧ share(R,S) ∧ base(F,U) ∧ structure(S,Y)
    ∧ goal(F,R) ∧ formalism(F,G) ∧ grammar(G,Z) ∧ unification(U)
```

Note that a large number of nouns in English and related languages are of the *relational* type, i.e. they denote relationships. One standard way to represent them in logic is by means of predicates of the appropriate arity, i.e. the same way we will normally represent verbs. Thus two-place predicates like `representation(R,L)`, `formalism(F,G)` etc. express the facts that we are talking about the representation *of* languages, formalisms *for* grammars etc. Note that none of these relations can be represented by the standard IR operators (such as *'adjacent'* etc.) since one of the properties of natural language is that the distance between functionally linked words can (often) be *arbitrarily* long.

The expression 3a is, however, not a logical sentence, i.e. it has no truth value and cannot, as such, be treated as an axiom. But in a retrieval context it is justifiable to say that, whatever is referred to by means of a full noun phrase, is asserted to exist. We can therefore translate everything that is explicitly referred to in a text by means of a full noun phrase into an existentially quantified statement, i.e. we can apply *"existential closure"* to formulae like 3a and get:

3b) $\exists$ R,L,S,Y,F,G,Z,U: representation(R,L) $\land$ language(L) $\land$ share(R,S) $\land$ base(F,U)
$\land$ structure(S,Y) $\land$ goal(F,R) $\land$ formalism(F,G) $\land$ grammar(G,Z) $\land$ unification(U)

In order to have back-pointers to the original documents we add, to each individual axiom, document number and, for passage retrieval, fragment (i.e. sentence, title, or caption) number as additional non-logical constants. If we use a subset of First Order Logic for which efficient proof procedures are known, such as Horn Clause Logic, we will get

3c) representation(sk-1,sk-2)/1/3, language(sk-2)/1/3, share(sk-1,sk-3)/1/3,
base(sk-5,sk-8)/1/3, structure(sk-3,sk-4)/1/3, goal(sk-5,sk-1)/1/3,
formalism(sk-5,sk-6)/1/3, grammar(sk-6,sk-7)/1/3, unification(sk-8)/1/3

meaning that these axioms were all derived from fragment number *1* in document number *3*. Existential quantification is encoded in HCL by means of so-called Skolem-constants, i.e. they are given arbitrary but unique names, here of the form 'sk-N' with *N* an integer. All variables are (implicitly) universally quantified.

Passage retrieval can now be interpreted as proving queries over the logical data base derived from the documents. If we apply a standard proof technique like refutation resolution, a query must be translated into the clausal form of its negation, and

**4) ?- Structure sharing representations of languages**

would then become

4a) ?- representation(R,L)/S/D, language(L)/S/D, share(R,S)/S/D, structure(S,Y)/S/D.

Note, first, that this query means that we want to prove the theorem over axioms that were all derived from the *same* sentence in the *same* document. Note also that the case where a query is more specific than a relevant document (the normal case) is taken care of automatically by this proof procedure, i.e. theorem 4a can be proved *directly* over the axiom system 3c.

## 5. The choice of a suitable ontology

One point that previous approaches to the problem of formalising the semantics of natural language have paid insufficient attention to is that of the *ontology*, or conceptualisation. The question of what kind of *objects*, and what kind of *relationships* between these objects, we assume to exist in the world has a direct bearing on the model to use, and therefore on the definition of the language. The first problem to address is that of representing relationships in logic. There are two basic approaches to this: The ''ordered argument'' approach, and the ''thematic role'' approach (cf. [5]:84ff.). The ordered argument approach assumes that there exists a finite number of roles that objects can play in relationships. These roles are *implicitly encoded* by means of the fixed argument positions of predicates. A sentence like

5) **John gave Mary an apple**

might thus be represented as

```
5a) gave(john,sk-1,mary).  apple(sk-1).
```

This is the approach taken in all the examples so far. However, there are many more roles expressed in natural language than those used so far, and for all of them we need to create additional argument positions. Take example

6) **On Tuesday, John furtively gave Mary an apple in the courtyard**

where we express information about the circumstances of the event, viz. about the *time*, the *location* and the *manner* in which the action was performed. We would thus have to extend the predicate pattern by additional argument positions and write

```
7) give(john,sk-1,mary,tuesday1,courtyard1,furtive).    apple(sk-1).
```

In natural language, there is a rich set of other circumstantial modifiers, for the *cause* and/or for the *reason* of an event ('...*because* he loves her'), for the goal/purpose (''*in order to* impress her''), or even for highly complex roles such as a concessive circumstance ('...*although* he was told not to'). It seems entirely unclear how many such roles ought to be expressed. There is growing consensus that it is not even a finite set.

Since the ordered argument model seems to fail to account for these empirical facts of language the alternative account, the thematic role account, has gained much popularity. This approach assumes a potentially infinite set of event roles. They are *explicitly encoded* as predicate names. The corresponding predicates range over one object each, and the event itself. What was an *n*-place predicate under the ordered argument account thus becomes a set of *n+1* two-place predicates. 6 would yield a logical representation like

```
7a) eventuality(give,sk-5). time(tuesday1,sk-5). agent(john,sk-5).
    location(courtyard1,sk-5). aff_ent(sk-1,sk-5). manner(furtive,sk-5).
    goal(mary,sk-5). object(sk-1,apple).
```

where the Skolem constant `sk-5` denotes the eventuality (the term ''eventuality'' subsumes actions, events, and states). If additional information concerning other event roles becomes available, the set of predicate names is monotonically extended (`reason`, `concessive` etc.), that is, we need never commit ourselves to a fixed set of roles.

It is sometimes assumed that a commitment for one of these two approaches is a matter of technical convenience. It is worth pointing out that this is not the case. Consider, for instance, the case of action modification (expressed mainly through manner adverbials): By translating 6 as 7 we implicitly assert, by the very act of creating argument positions for time and place, that these attributes are equally fundamental to a goal-directed action as are agent, affected entity, and goal. In particular, by creating an argument position we are forced, under the Horn Clause Logic model assumed here, to specify that the corresponding variable is either universally or existentially quantified. We cannot leave its quantificational status unspecified. However, in natural language we seem to allow unspecified attributive values. While utterance 8 is syntactically well-formed (no time and place is specified), 9 is not (no affected entity is specified):

8) **John gives Mary apples**
9) ***John gave Mary on Tuesday in the courtyard**

It can thus be seen that agent, affected entity, and goal are *obligatory* roles while time and place are not. Note that generic sentences like 8 do *not* quantify over points in time or locations, not even implicitly (e.g. by assuming default values). Utterance 8 asserts that it is one of John's habits to give Mary apples, without any indication of place or time. In particular, utterance 8 is neither synonymous with ''For *any* given point in time, John will be seen giving Mary apples'' nor with ''For *some* point in time...''. Since the interpretation for the clausal representation enforces, under resolution refutation, a binding for each argument value of a predicate (either referential, existential or universal), we could not represent 8 under the ordered argument scheme without an unwarranted commitment to the values of attributes like ''place'' and ''time'' (and any other attributes encoded as argument positions).

Under the thematic role scheme (axiom system 7a above), on the other hand, we treat all attributes as equally peripheral and dispensable. We could perfectly well drop the roles for agent, affected entity and goal without the resulting set of axioms becoming incoherent. However, any rendering of it in natural language would result in an ill-formed utterance, such as 9. Moreover, the vital distinction between obligatory and optional information is not made, either. We believe that this shows that neither of the representation schemes can be used in pure form.

There are a number of syntactic reasons that convince us that an *intermediate position* must be found. Most substantial among them is the basic syntactic distinction between *complements*, which are obligatory (at least in declarative main sentences; cf. [6]:481) and *adjuncts*, which are *optional*. More specifically, we argue that the former correspond to the semantic category of *participants* in eventualities, and the latter to *circumstances*. The most appropriate way to map such constituents into a logical representation is therefore to use fixed argument positions in a complex *main predicate* for the values derived from complements and the (equally obligatory) governing subject (this is the ordered argument component of the compromise), while all others are represented as *auxiliary predicates* (this is the thematic role component). Main and auxiliary predicates are linked through one additional argument for the eventuality identifier. This intermediate position corresponds materially to Davidson's original approach ([7]). Later objections to it (also by Davidson himself), based primarily on the possibility to report in a non-contradictory manner on impossible situations ([8]:87 ff., [5]:93ff.), fail to convince us.

For current purposes we may suppose that there are three participants in English, viz. **Agent, Affected-Entity, and Goal**. For a sentence like 6 above, with a prototypical bi-transitive main verb with clear complements, this might give the following set of axioms

```
7b) action(sk-5,give,john,sk-1,mary). object(sk-1,apple).
    location(courtyard1,sk-5). time(tuesday1,sk-5). manner(furtive,sk-5).
```

If, in a last step, we define sortal restrictions on the possible values that the arguments can take (e.g. `+animateness` for **Agent**) and make sure that for each eventuality there can be only *one* main predicate, the resulting representation scheme amounts to a *case frame representation* for the central propositions of sentences with a sound logical foundation.

In this section we move from the problem of the required *number* of participants in eventualities to the topic of their *semantic characteristics*. The ontological intuition behind the use of a fixed number of argument values is that the fillers of a given thematic role have certain semantic characteristics in common, *irrespective* of what concrete predicate is used to denote the relationship. If we continue to use the thematic roles most commonly assumed to exist, viz. **Agent, Affected-Entity**, and **Goal**, and an appropriate predicate scheme of the following form

```
action(Id,Action,Agent,Aff-Ent,Goal)
```

these thematic roles would allow us then to postulate general inference rules like

```
∀ I₁,P₁,A₁,E₁,G₁,I₂,P₂,E₂ action(I₁,P₁,A₁,E₁,G₁) ∧ action(I₂,P₂,E₁,E₂,G₁)
   → ∃ I₃: action(I₃,P₂,A₁,E₂,G₁) with P₁ ≠ P₂, E₁ ≠ E₂, I₁ ≠ I₂ ≠ I₃
```

that is, "If Agent $A_1$ performs some Action $P_1$ to Affected Entity $E_1$ with Goal $G_1$, and $E_1$, as Agent, performs some other Action $P_2$ to Affected Entity $E_2$ with the same Goal $G_1$, then Agent $A_1$ effectively himself performs Action $P_2$ to Affected Entity $E_2$ with Goal $G_1$".

It would certainly be desirable to have this kind of completely general inference rules but there is so far very little consensus on the uniform semantic characteristics of different thematic roles that would make such rules possible. Somers, at the end of an exhaustive investigation of the literature ([9]), came to the conclusion that it is *not* possible to find a small and closed set of thematic roles which capture all the semantically relevant information about role fillers. He distilled off four very general *"inner roles"*, modelled (and named) after the prototypical processes of movement, viz. ''source'', ''path'', ''goal'', and ''local'' (± affected entity), and then defined a number of *parameters* which combine with the inner roles to give concrete participant descriptions. These parameters correspond to very general types of eventualities, of which Somers suggested six, beginning with straightforward movement in space and time (''locative'' and ''temporal''), over actions (with an active agent: ''active'') and processes (without an active agent: ''objective''), to immaterial changes of (psychological or legal) eventualities (''dative'') and finally eventualities without any agent (weather verbs etc.; ''ambient''). Consequently, semantic generalisations are allowed only *within* a given type of eventuality while the same inner role (say, ''source'') in two *different* types of eventualities (say, ''objective'' and ''dative'') has no common semantic characteristics.

This explanation of the facts is very attractive from a computational perspective ([10]), in at least two respects.

## 6. A mixed-level representation

*First*, inferential relationships in terms of *types* of eventuality are at least as useful as completely general inference rules operating over true deep cases. Using a predicate scheme of the following form

```
Parameter(Id,Eventuality,Source,Path,Goal,Local)
```

we can formulate rules like

$\forall$ $I_1, E_1, S_1, P_1, G_1, L_1, I_2, E_2, P_2, G_2$: `locative(`$I_1, E_1, S_1, P_1, G_1, L_1$`)` $\wedge$
`locative(`$I_2, E_2, S_1, P_2, G_1, L_1$`)` $\rightarrow$ $\exists$ $I_3, E_3, P_3$: `locative(`$I_3, E_3, S_1, P_3, G_2, L_1$`)`
*with:* $E_1 \rangle E_2 \rangle E_3$ (non-overlapping temporal sequence) $\wedge$ [[ $E_3$ ]] $\supset$ [[ $E_1$ ]] $\wedge$ [[ $E_3$ ]] $\supset$ [[ $E_2$ ]]

Provided we know that both ''roll'' and ''fall'' are *moving*-actions (parameter *locative*) we can now perform the inference

```
Since the ball rolled from the center of the table
to its edge and fell from there to the floor
it will have moved (in an unspecified manner)
from the center of the table to the floor
```

Significantly, such inferences are on the right level of granularity for the purposes of natural language understanding.

*Second*, it is now relatively easy to recognise the ''inner roles'' for a given verb phrase as we no longer require that there must be common semantic characteristics to *all* fillers of a given inner role, irrespective of the type of eventuality. We can therefore apply a fairly shallow and straightforward mapping from grammatical structure to inner role structure, using mainly the subcategorisation information for the given verb type. In many cases, the inner roles are linguistically realised by means of particularly obvious *spatial expressions*, even when used for types of eventualities that have nothing to do with space at all, as in ''to translate *from ... to*'' (probably a reflex of the historical development of language). The mapping may be different for the various types of eventualities but for each given type they must be the same, or else the assignment of roles will be useless for the purpose of inferences. In fact, this observation can be used to inductively infer what types of eventualities should be distinguished ([6]:481): Those verbs whose role frame allows the same inferences denote the same type of eventuality. To actually infer types of eventualities that way, presumably on the basis of corpus material, would require a massive investment in time and resources which, to the best of our knowledge, has never been made. This is why we use, for the time being, a very small subset of eventuality types (after [11] and [12]), viz. ''states'', ''processes'' (''actions'' where the agent is active), and ''events'' (''performances'' with an active agent). In view of the data presented above it seems fairly clear that a parametrised role concept is much easier to implement than the elusive fully general ''deep case'' approach (and probably it is more useful in actual applications).

Once we have determined what roles and what eventuality parameters we have to distinguish, and how syntactic structures relate to them, we must decide how many *types of modifiers* should be used, i.e. how many additional expressions of circumstances must be introduced. Expressions of time and location are the most obvious ones and manner, cause, and reason may also be uncontested candidates, but what about the innumerably many other ways to express additional information about events and actions? Are there hard criteria telling us what kind of modification should be dignified with its own role predicate? The linguistic evidence strongly suggest the appropriateness of a *layered representation system*: Instead of trying to find a closed set of modifiers it seems reasonable to claim that there is a small, finite, core set of fairly general (and easy to recognise) modifiers, and an outer layer of (arbitrarily specific) additional circumstance descriptions, whose number is potentially infinite.

The set of core modifiers we use so far consists of just **purpose**, **method**, **tool**, **beneficiary**, and **manner**. For all modifiers that we cannot analyse in these terms we resort to a *lower level of abstraction* and represent them in the logic as *themselves*, i.e. they become non-logical constants. In 10

**10) On Tuesday, John gave Mary a nice computer table against her will**

we do, for instance, not know whether ''computer table'' means ''table next to the computer'', ''table on which to put a computer'', ''table designed by computer'', or any of a number of other possible readings, and it is equally unclear what kinds of semantic relationships are encoded in the prepositional phrase ''against her will''. We therefore turn the unanalysed preposition and the implicit ''of'' in the personal pronoun into non-logical constants, and for the nominal compound we create an artificial constant (`by_with_for`) encoding the unanalysed relationship between its constituent parts. That way we get a *mixed-level representation*, combining expressions on *three levels of abstraction*:

1. **Expressions of level 1:** A fixed number of general, obligatory and unique thematic role fillers

2. **Expressions of level 2:** A fixed number of general but neither obligatory nor unique modifier predicates

3. **Expressions of level 3:** An unlimited number of arbitrarily specific, optional and non-unique circumstantial descriptions

An appropriate representation of utterance 10 would therefore be

```
10a) locative(give,sk-1,john,sk-3,mary). object(table,sk-3).
     object(computer,sk-2). object(will,sk-4).  property(nice,sk-3).
     time(sk-1,tuesday).
     circumstance(by_with_for,sk-3,sk-2). circumstance(against,sk-1,sk-4).
     circumstance(of,sk-4,mary).
```

with expressions of level 1 at the top, followed by expressions of levels 2 and 3. By using different levels of semantic granularity in the knowledge representation, we can refine our knowledge in a fully incremental fashion

- We always represent knowledge at the most general level possible at the given point in time but allow for arbitrarily specific entries.

- If we should later discover more general entailments between some of these specific entries, we assert appropriate rules, i.e. *meaning postulates*, found empirically and added incrementally.

- The more information we gather, the denser the network of such entailments will become, without any need to restructure the knowledge base.

In the given application context of passage retrieval, meaning postulates are particularly important to increase recall. Consider a passage

**11) Natural language question answering systems**

where we speak about *systems* that perform certain actions by means of natural language, with the logical representation:

```
property(natural,sk-28)/1/11. object(system,sk-30)/1/11. object(language,sk-28)/1/11.
circumstance(by_with_for,sk-30,sk-28)/1/11. object(question,sk-29)/1/11.
eventuality(answer,sk-31,sk-30,sk-29)/1/11.
```

But consider now the query

**12) Natural language questions**

which is obviously about *questions* phrased in natural language. Nevertheless we would definitely want query 12 to retrieve passage 11. We can increase recall of a retrieval system suitably by using meaning postulates such as

```
circumstance(by_with_for,O1,O2)
   <-  eventuality(AType,Ev,Ag,O1), circumstance(by_with_for,Ag,O2).
```

which will also allow us to retrieve document 11 through query 12.

## 7. Variable-Depth evaluation

Unfortunately, the use of a mixed-level representation does not come for free. The lack of a known and fixed level of generality must be compensated for by a large number of meaning postulates, and these tend to get very detailed, i.e. their branching factor is high. On top of that, in most application contexts we will have to use inheritance hierarchies. If we use meaning postulates and inheritance relationships *whenever* they are applicable we will run into insurmountable problems with the size of the search space. We need a mechanism that controls the search procedure suitably. This is where *variable-depth evaluation* becomes useful ([13], [14]).

We can distinguish two types of criteria that may control the search procedure: *External criteria* (such as the maximum number of passages to be retrieved), and *internal criteria* determining the resources to be used for a given proof step. External criteria are straightforward to implement: If a relatively shallow level of evaluation has already produced a large number of results (e.g. relevant passages), it is better to stop the proof at this level and present the results to the user who might then be able to re-phrase the original query and make it more specific. External criteria stop the proof when it is, as it were, *too* successful. Internal criteria influence the proof process when it is *not enough* successful, and steer it towards more promising branches of the search space. The single most important internal criteria are probably the *total number of inferences* allowed for the proof of a given term, the *type of the rules* to be used, and the *weights of individual rules* (indicating, for instance, their reliability or general usefulness).

In LogDoc, we use a combination of external and internal criteria. The different computational costs of using various types of rules, and also their different usefulness as determined by preliminary experiments, suggested the following strategy:

1. if we have found more than M passages using only *direct* matches of thematic role relations, we do not use any of the meaning postulates

2. if we have found fewer than N (N < M) passages we begin to use meaning postulates, first those on level 2 (defining modifiers), then those on level 3 (defining circumstantial descriptions)

3. if we have found fewer than O (O < N) passages so far, we also try inheritance hierarchies

We found that a reasonable weighting of these values is M:N:O = 3:2:1. The absolute threshold values used so far are very low (M=15, N=10, O=5), due to the very small size of the sample of documents used so far. However, we seem to get a reasonable ordering of passages.

## 8. Implementation and Results

LogDoc is implemented as a fully functional prototype over a small set of bibliographic entries used as documents (on a previous version, see also [15]). While the two main concepts described so far, viz. mixed-level representation and variable-depth evaluation, are not specifically linguistic in character and generalize to more complex examples in other fields of AI, the machinery to make natural language documents and queries amenable to this analytic approach, is eminently linguistic. We will now describe the main linguistic components of LogDoc and give some indication as to the system's performance and limitations.

### 8.1 *The syntactic background*

LogDoc was designed to work with any kind of syntactic theory. Its syntactic and semantic components are kept strictly separate, and their only channel of communication is used to pipe *functional structures* (f-structures) from the former to the latter. The only requirement for the syntactic module is therefore to produce f-structures representing the fundamental grammatical functions of the constituents in a sentence (such as: the sentential main verbal predicate; verbal complements like subject, direct object, indirect object; nominal, verbal and sentential pre- and post-modifiers). It turned out that, for the fairly modest grammar needed for our experiments, the *Definite Clause Grammar* formalism was sufficient but any of the syntactic frameworks popular today could be used with (almost) equal ease (other, more sophisticated, unification based frameworks like GPSG or HPSG; transformational frameworks like Government and Binding or Minimalist Theory; dependency based frameworks).

It is, in fact, much more relevant what *object language* is to be used than in what *syntactic theory* the grammar of this language is to be cast. It is, for instance, much easier to create f-structures for languages with relatively strict word order like English than for languages such as German. Nevertheless the modularization of syntax and semantics allowed us to use the system, which was originally developed for bibliographic entries in English, also for the subject lines in administrative letters in French, with negligible porting costs (apart, evidently, from the considerable costs of developing the French morphology, lexicon and grammar components).

### 8.2 *The parser*

The parser plays a pivotal part in any NLP system, and LogDoc makes no exception here. The particular uses to which syntactic analyses are put in LogDoc had, however, a considerable influence on the kind of parser we developed. First, in neither of the two experiments performed we could expect the input to be well-formed *sentences* in any but a few cases (neither titles of scientific articles nor subject lines of letters are normally complete sentences). The parser would therefore have to be able to cope with individual phrases. Second, it was clear from the outset that the resources at our disposal would not allow us to develop a grammar that covered more than a small percentage of the phrases occurring in the input. For both reasons we needed a technique for *partial* syntactic analyses. This meant that we had to use a parser that analyses as much of the linguistic input as it can and returns the rest unanalysed. The prime example of a parser that shows this behaviour is the so-called *"chart parser"*. A chart parser builds up all possible syntactic analyses simultaneously, and keeps all the syntax structures in one complex data structure (i.e. the chart). Furthermore, if we make the parser work *bottom-up* and *breadth-first* and do not make any assumptions about final categories, the parser will have accumulated all partial parses of the input once it finds out that it cannot analyse the entire input as one coherent phrase. We can then extract the *largest meaningful* parts of syntax structure from the chart, and hand them over to the semantics component for further processing. It is, incidentally, not quite obvious what we are to take as the "the largest meaningful part". Contrary to what we expected the *longest* constituent recognized by the system sometimes is *not* part of the intended analysis (as judged by humans). Every so often it happens that a shorter stretch of analysis is better. We have not yet found a satisfactory explanation for this observation.

One major problem in NLP systems (arguably *the* main problem) is the ambiguity of natural language. Ambiguity comes in different forms but one of them is particularly insidious: Structural ambiguity. Structural ambiguity occurs whenever the same sequence of word forms can by analyzed in different ways and get, accordingly, different syntactic structures. The problem of ambiguity actually consists of two sub-problems: On the one hand, in most cases of ambiguity *one reading* is "obviously" the *intended* one, but it is unclear how to determine which one (as seen below). The fact that we humans find the answer often extremely obvious makes it likely that we use a considerable amount of "common sense" world knowledge when we analyze such expressions. Unfortunately, it is exactly this kind of knowledge that is hardest to model on the computer. On the other hand, natural language constructions are often *genuinely ambiguous*, i.e. human readers agree that no strongly preferred reading exists. What is a language processing system supposed to do in this situation? We will consider these two sub-problems in turn.

Since each syntactic structure of am ambiguous string has a potentially different meaning ambiguities are a major problem whenever we need to process the semantics of natural language utterances. Consider, for instance, the (far from exotic) noun phrase

**13) A new characterization of attachment preferences in English**

Even in the very simple-minded grammar of English used in LogDoc, it has no fewer than 38 different syntactic analyses, and therefore 38 potentially different meanings. The *intended* reading requires that the prepositional phrase "in English" is attached, to "attachment preferences" so that we get

```
13a:   object(english,sk-49)/13. object(attachment,sk-50,sk-51,sk-52)/13.
       object(characterization,sk-53,sk-54,sk-55)/13.
       object(preference,sk-55,sk-56,sk-50)/13.
       property(new,sk-53)/13. relationship(in,sk-55,sk-49)/13.
```

where `relationship(in,sk-55,sk-49)/13` expresses the fact that it is the *preference*, `sk-55`, that is *in* English, `sk-49` (i.e. "this preference occurs" or "is observed", in the English language). For human readers it is quite obvious that this is the intended reading of this phrase, and many people will, in fact, be under the impression that there are no other readings available. However, a syntactic analysis with a different attachment of the prepositional phrase (viz. to ''characterization'') is, in purely syntactic terms, perfectly possible, with the resulting meaning that it is the *characterization* which is in English (say, the characterization is *phrased in English*). Of the 36 other syntactic analyses 20 happen to by synonymous with one of the other analyses but this still leaves us with 18 different meanings! And slightly longer phrases can easily receive hundreds of different analyses with several dozens of different meanings, particularly so when a phrase is analysed with a grammar that is larger than ours (the higher the coverage of a grammar is the more numerous the spurious analyses tend to be).

The importance of this problem is, in principle, acknowledged in the literature on natural language based IR (for instance [16]:98) but it seems that most authors in the IR community do not see in it the kind of massive problem that it really is, and they assume that a suitable solution can be found somehow. However, anybody active in NLP knows that this problem is an absolutely central question. There is general agreement that any solution will have to use some kind of preference scheme that imposes a *plausibility ordering* on the possible syntactic analyses. Two types of (not mutually exclusive) ordering schemes are normally used: General rules that give preference to trees with a certain *overall geometry*, and specific rules that compute overall preferences on the basis of the likelihood of occurrence of *local patterns of word categories*. Under the first approach, wide and shallow trees could, for instance, be preferred over narrow and deep ones, or vice versa. Under the second approach, the analysis '((adjective - noun) - noun)' could be preferred over '(adjective - (noun - noun))', or the other way round. Our experiments with LogDoc convinced us that the first of these ordering schemes can be used unchanged, that the second scheme must be modified in two respects to be really useful, and that a third, additional, scheme is required that operates on an intermediate level of syntax structures, viz. *sub-sentential phrases*. We will discuss these three types of preference rules in turn.

*First*, we look at general preference rules. A number of them have been suggested in the literature but the two best known are the ''Principle of Right Association'' and the ''Principle of Minimal Attachment''. The first one (also called ''Low Right Attachment'', ''Late Closure'', or ''Local Association'', originally due to [17], modified by Frazier and Fodor in [18]) says that any new constituent will be attached as deep and far to the right as possible in the syntax tree built up so far. This principle would favour the first reading of example 13 above (13a). The second principle, the ''Principle of Minimal Attachment'', says that a new constituent should be attached to a growing tree in a way that requires the creation of as few new (non-terminal) nodes as possible. Again, the first reading of example 13 would be preferred. Unfortunately, the two principles do not always agree. Consider

**14) The operator tested the programs on the system**

This sentence has among its readings

**14a) The operator tested the programs *which are* on the system**

**14b) The operator tested the programs *by means of* the system**

Under reading 14a, the prepositional phrase ''on the system'' modifies just the noun ''the programs'', under reading 14b, however, it modifies the verb phrase ''tested the programs''. Here, the Principle of Right Association would prefer the first reading with a syntax analysis tree comprising 19 non-terminal nodes, the Principle of Minimal Attachment, however, the second (18 nodes). Intuition clearly tells us that both readings are possible but a slight preference seems to be given to the first one.

We have therefore to, first, *combine* these two general rules so that they interact properly, and we must, *second*, make sure that the Principle of Right Association gets a slightly greater influence than the Principle of Minimal Attachment. Each general rule can be implemented by a method that computes the overall preference value from the preference values of sub-structures in a way that encodes one particular property of the tree (e.g. number of nodes, number of embeddings). These two general rules are encoded in LogDoc as follows:

1. We use a simplified form of the Principle of Right Association which prefers deeper trees over shallow trees irrespective of the *place* where additional branches are attached. This principle turned out to be more reliable than the Principle of Right Association proper. We can implement this version of the rule by assigning to a given structure a *reward* for every additional level of embedding.

2. The Principle of Minimal Attachment is implemented by assigning a *penalty* to every additional node. For our grammar (which tends to create a large number of nodes) this penalty must be very high.

The right balance between the influence of the two principles must be determined. The question is: When is the benefit of creating a deeper tree outweighed by the cost of the additional nodes needed?

We get the right behaviour if the overall preference for a given constituent α, viz. $V_\alpha$ is computed by the following, very simple, function

$$V_\alpha = \frac{\sum_{i=1}^{i=n} V_i}{\text{Rew}} + \text{Pen}$$

with **Pen** the constant node penalty, **Rew** the constant level reward, and $V_i$ the preferences of the constituent sub-structures of α (recursively computed from *their* respective sub-structures). Note that a preference value is the *higher* the *less* likely the corresponding reading is considered to be. The values of *Rew* and *Pen* have been empirically determined to be, for English **Rew** = 2.25 and **Pen** =15. This function together with the parameter values give approximately the right distribution of preferences, even in the absence of specific preference values (see next section).

We now turn to the *second* type of preference rules, viz. those taking into account local patterns of word categories. This is the type of rule that should tell us whether, for instance, the sequence '((adjective - noun) - noun)' is to be preferred over '(adjective - (noun - noun))', or the other way round. A particularly striking example where such rules would be useful are sentences such as 14 above, repeated here

**11) Natural language question answering systems**

On the basis of the general rules *alone* this sentence would get an analysis corresponding to the bracketing 11b below, which is clearly inappropriate (this analysis is synonymous with ''natural systems answering questions concerning language''). However, the intended reading requires bracketing 11c:

```
11b) (Natural (((language question) answering) systems))
```

```
11c) ((Natural language) ((question answering) systems))
```

However, a suitable preference rule operating on the level of word categories would have to prefer the bracketing

```
((Adj-N)-((N-Gerund)-N))
```

over any other, which would give the wrong prediction for

**15) Specific preference rules preferring sentences**

since, in this case, the intended reading requires the bracketing

```
15a) (((Specific (preference rules)) preferring) sentence)
```

The reason why 11b is the intended reading turns out to be much simpler: The sequences of *lexical items* ''natural language'' and ''question answering systems'', respectively, almost always belong together, at least in the context of Artificial Intelligence, i.e. they are so-called *set phrases*. This is the first respect in which we have to modify the second type of preference rule. We allow highly specific preference rules where each set phrase contributes an *additional reward* (of varying amount, to be determined empirically) towards the preference of the entire constituent. It was found that such additional rewards are best designed as *subtractive* correction constants. The function for the computation of preferences is therefore modified as follows:

$$V_\alpha = \frac{\sum_{i=1}^{i=n} V_i - Spec_\alpha}{\mathbf{Rew}} + \mathbf{Pen}$$

with $Spec_\alpha$ the value of the specific preference value of constituent $\alpha$. It is now possible to fine tune the cohesion of set phrases. The higher the value of a correction constant, the more strongly will a phrase stick together.

While set phrases are easy to handle, they are not very numerous, and each specific preference value takes care of only one set phrase. They would *not* apply to the following examples

**16) answering machines**
**17) implementing languages**
**18) backtracking problems**

Again, all three phrases would get, through the interaction of the two general preference rules, as most highly ranked analysis that of 16 (i.e. ''languages that implement'' and ''problems that backtrack''). Again, they all have the same sequence of word categories, so the second type of preference rule would not be able to differentiate between them. Yet we know, intuitively, that the intended readings are, for example 16 ''machines that answer (something)'' (not ''the action of (someone) giving answers to machines''), for example 17 ''the action of (someone) implementing languages'' (not ''languages that implement (something)'') and, for 18, ''problems connected with backtracking'' (not ''problems that backtrack'').

In order to be able to get the correct readings we need information about the *semantic types* of the words involved. Machines are *active objects*, i.e. objects that can initiate and perform actions (e.g. answer to something) while languages are *passive objects* that hardly ever *do* things. On the other hand you can implement languages but hardly backtrack problems. These examples show that local effects which are governed by the *sort of the objects* denoted by the lexical items involved, are often much stronger than the global effects induced by the *geometry of the syntax structures* and the local effects induced by sequences of word categories.

We can get the correct behaviour if we define the following three rules that increase the preference values of the intended analyses (by the value 80) *if* the head noun of the common noun phrase (`CNp`) is of the appropriate semantic type (i.e. either `active_object` or `passive_object`), and if the verb also belongs to the correct type (viz. `activity`, as opposed to `process`):

```
% "answering machine": "a machine that answers"
preference(cnp_verb,CNp-verb(Verb),intr,80)
           :-    head(CNp,noun(N)),
                 active_object(N),
                 activity(Verb).

% "interpreting queries": "someone interprets queries"
preference(gerund_np,CNp-verb(Verb),_-tr,80)
           :-    head(CNp,noun(N)),
                 passive_object(N),
                 activity(Verb).

% "backtracking problems": "problems with backtracking"
preference(gerund_as_cnp,cnp(_,[],verb(Verb),[],[],[],[])-CNp,80)
           :-    head(CNp,noun(N)),
                 passive_object(N).
```

If these type restrictions are not satisfied, no specific preferences values are used, and the analyses will be those induced by the general preference rules alone.

The three grammar rules that access these preference values (via the identifiers `cnp_verb, gerund_np, gerund_as_cnp`) are, in order:

```
cnp(V,cnp(V,adjp(V,[],cvp(V0,Verb,[],[],[],_,gerund)),CNp,[]),_)
        --->    verb(V0,Verb,Tr,Nr,gerund),
                cnp(V1,CNp,Nr),
                {preference(cnp_verb,CNp-Verb,Tr,Pf),
                combine([V0,V1],Pf,V)}.

np(V,np(V,[],vp(V1,CVp,Pp,ComplS,Nr,gerund)),sing)
        --->    vp(V1,vp(V1,CVp,Pp,ComplS,Nr,gerund),Nr,gerund),
                {CVp=cvp(_,verb(Vb),CNp,_,_,_,_),
                transitivity(Vb,Tr),
                preference(gerund_np,CNp-verb(Vb),Tr,Pf),
                combine([V1],Pf,V)}.

cnp(V,cnp(V,[],Verb,[],[],[],[]),sing)
        --->    verb(V0,Verb,_,_,gerund),
                {Verb = verb(VV),
                transitivity(VV,intr),
                preference(gerund_as_cnp,Verb,Pf),
                combine([V0],Pf,V)}.
```

The grammar formalism used, that of Definite Clause Grammars (DCG), allows us to integrate the preference machinery described above in a very straightforward manner with the syntactic components of the grammar proper: In grammar rules, the terms *outside* curly brackets on the right hand side define the constituents of the phrase given in the rule head, and their ordering, while terms *inside* curly brackets express additional, non-input-consuming, tests. It is those tests that control the preference machinery. Thus the predicate `combine/3`, implements the preference function given above by computing the total preference value (*V*) from the individual preference values of constituents (*V0, V1*) and the preference value for the given phrase type itself (*Pf*) as defined in the terms `preference/3`, and in addition it takes into account the number of nodes in the syntactic structure being built up, and the level of embeddings so far. The total preference value, *V*, is then used by the parser to dynamically prune the search space. No further provisions have to be made in the grammar to allow this kind of pruning as all settings (preference values for individual phrases, default preference value, node penalty, embedding reward) can be done centrally.

More difficult are those cases where the relevant words do not occur next to each other in the input string, like

**19) The operator translated the sentences with a computer**

which is structurally identical to example 14 above and has, of course, the same two syntactic analyses, corresponding to the paraphrases

**19a) The operator translated the sentences *which are connected with* a computer**

**19b) The operator translated the sentences *by means of a* a computer**

The combined effects of the two general preference rules would, again, give the first reading (19a) a higher overall preference, as in example 14. It is, however, clear that this is the far less probable reading. One idea to differentiate these sentences is to assume that verbs have certain ''expectancies'' concerning their arguments, i.e. they have a *case frame* whose slots must be filled with complements of the correct type. A verb such as ''translate'' has a case frame like

```
[ AGENT translate PATIENT TOOL]
```

which gives any analysis where all the case slots are filled a higher preference over those where some slots are left unfilled (such as TOOL, which remains unfilled under the analysis 19a).

This idea, which has a venerable history in linguistics (where type restrictions on fillers of case frame slots are known as ''selectional restrictions'') is easier to describe in general terms than to implement in a concrete system. Both the information on the case frames (complete with the types required for the slot fillers), for several thousand verbs at the very least, as well as for the types of potential complements is not readily available. In addition, many nouns (such as ''design'') and even adjectives also have such case frames associated with them. As no resources containing this kind of information were available to us at the beginning of the project we had to opt for a more limited approach which turned out to be surprisingly powerful: For around 20 particularly critical *sub-sentential syntactic structures* (i.e. above the level of word categories), specific preference values have been determined empirically. This is the *third* type of preference rule that we added to the two traditional ones. A detailed description of the individual preference dependencies is beyond the scope of this paper but, by way of example we can mention one rule, viz. the one that picks the intended reading of example 19. It states that the attachment of a prepositional phrase to a verb would get a lower preference value than its attachment to a common noun phrase. Again, in many cases the preference values in such rules are parametrised by the semantic type of one, or several, of the lexical items that instantiate the terminals (`active_object` vs. `passive_object`).

The preference machinery described so far will *rank* the analyses according to their overall preference values. But even if the ranking is determined correctly it cannot be assumed that the first, top ranking, analysis will always be the *only* intended reading. Quite often human readers will consider the differences between the first few top analyses so small as to be insignificant, i.e. the sentences are felt to be genuinely ambiguous. Sentence 14 above is a case in point. It is not a trivial problem to find a way to suppress all, and only, those analyses felt by human readers to be unavailable. We considered three algorithms:

1. choose the first *N* top ranking analyses
2. choose the analyses with a preference value of at least *N*% of the preference value of the top ranking analysis
3. choose those of the top ranking analyses whose preference values do not differ by more than *N*% of each other

The first method soon turned out to be at variance with intuition. If *N* was assumed to be *one* even genuinely ambiguous sentences would get just one reading. To assume that *N* is *two* or more, is even less reasonable as even genuinely *un*ambiguous sentences would now (virtually always) get several readings. But the second method, too, could in many cases not satisfy. The human reader does not seem to consider the *entire* range of preferences (from least preferred to most preferred reading) when determining how many of the top ranking readings should be assumed valid. It seems that humans consider only the *local* context of preferences, i.e. how closely spaced the individual preference values are. For this reason it was the third approach that turned out to be most suitable. It forms *clusters* of values by computing the *coefficients of successive preference values*. If the coefficient is below the (empirically determined) threshold of 0.897 the values belong to the same cluster. Only the first, top ranking, cluster of readings is taken.

Filtering out ambiguous readings at the end of the parsing procedure is necessary but it obviously does not improve the efficiency of the procedure, quite on the contrary. In order to get acceptable run-time behaviour we must add an *agenda mechanism* which makes sure the highest ranking hypotheses are pursued first. In addition, we also want to dynamically *prune the search space* during parsing. Both goals can be reached by ordering, for a given constituent, the newly created inactive edges of the chart parser according to their preference values and keeping only those with the best preference values. All other edges are discarded right away. This amounts to using a variant of *n-best search*, and it turned out to be advantageous to set *n*=1. Note that in many cases there are *several* edges with exactly the *same* preference value. This situation occurs whenever the parser runs into an ambiguity but does not have available any disambiguating information. In such cases it must assign the same preference value to the different readings. This means that even an n-best search with n=1 does not just pick one analysis (the first ranking one) but several (the first ranking ones).

This rather radical pruning procedure reduces parsing time drastically (by a factor of around 13) yet does not seem to result in a significant loss of correct solutions (a very similar approach is taken by [19]). Note also that the filtering stage described above is not made redundant by the pruning procedure as pruning lets through a fair number of parses that do not make the mark after all.

In approximately 80% of the examples in our experiments the combination of general and specific preference rules, combined with the filtering and pruning procedures just described, managed to determine the intended reading(s) of phrases and sentences. However, it became clear that further performance improvements could not be achieved by fine-tuning the preference values empirically (i.e. by hand) but that we will need to *train* the system to automatically acquire the optimal settings from syntactically analysed texts (so-called tree-banks).

*8.3 **The semantic component***

Once the syntax structure of an input string has been computed it must be transformed into HCL. The main three problems in this connection are:

1. subsentential syntactic analyses
2. partial syntactic analyses
3. unresolved ambiguities

The first problem is the result of the fact that input strings in our application domain are, in many cases, not complete sentences but subsentential phrases such as noun phrases or prepositional phrases. The semantic module of LogDoc must therefore be able to translate such subsentential phrases into logical forms. The second problem is that the limited coverage of the grammar will often not allow the parser to fully analyse the input string even if it is a complete sentence, and we will get several partial analyses (subsentential phrases or even individual words). They, too, must be translated into logic. Third, despite all the energy spent on filtering out ambiguities we will very often end up with several analyses, which must be considered equally probable. We must somehow translate all of them into logic.

The first two problems can be solved using the same method. What we need is a procedure which is capable of translating *any* well-formed syntactic structure, as small as it may be, into a well-formed, i.e. *interpretable*, logical expression. That this is not a trivial task becomes obvious if one tries to come up with interpretable logical forms for individual words, say, ''every'' or ''man'', or for phrases such as ''owns a dog''. Not only have the logical forms for these phrases to be interpretable, they must also be of a form that allows us to combine the logical form of ''every'' with that of ''man'' to get a third logical form that, in its turn, can be combined with the logical form of ''owns a dog'' to get the logical form of a complete sentence. What we need, in other words, is a *compositional semantics* of natural language which allows us to compute the meaning of phrases *alone* from the meaning of their constituent parts and from the way these parts are combined (i.e. the syntactic structure of the phrase). Luckily there are only relatively few constructions in natural language whose semantics is non-compositional, mainly idioms like ''to bite the dust'' where the meaning of the whole can, in fact, *not* be derived from the meaning of its constituent parts. This holds true even for languages like English where an above average number of idiom-like phrases is used (e.g. phrasal verbs such as ''to come round''). How the semantics of the other, compositional, constructions of natural language should look like was considered a deep problem until the late sixties.

The classical solution of this problem is due to Richard Montague. He showed that a compositional semantics for natural language is possible through what became known as *Montague semantics* ([20], [21]). The basic idea of this theory is simple in principle, and very difficult to implement in practice for any non-trivial fragment of a natural language. Montague showed that you can express the meaning of *every* type of phrase, from the individual word up the whole sentences, in a way that allows the computation of the meaning of any well-formed syntactic combination of such phrases through *functional application*. Thus, if we have a syntax structure like `sent(np(NP),vp(Vp))` (for, say, "Peter smokes") the meaning of the top constituent `sent` can be computed by taking the logical formula representing the meaning of `Vp` ("smokes´") as *functor* to be applied to the formula representing the meaning of `Np` ("peter´") as its *argument*. If we represent the meaning of a syntactic expression *T* as *T\** we would thus get

`Sent* = Vp*(Np*)`

This makes for translation rules which are so simple that they can easily be merged with the grammar rules themselves. For the above example the grammar rule

`sent(sent(np(NP),vp(Vp)),`**`TVp(TNp)`**`)  →  np(Np,TNp),   vp(Vp,TVp).`

thus defines syntax *and* semantics of a sentence.

However, this simplicity comes at a price. First, all constituents of the same *syntactic* type must be given the same *semantic* type also. The reason is that functional application will have to be used in *exactly* the same way to combine different constituents of the same syntactic type (e.g. "Np") with other constituents of a given type (e.g. "Vp") to give a higher constituent (e.g. "Sent"). Thus the semantic type of a proper name (such as "John") must get the same type as a quantifying phrase (such as "every man"), viz. "Np". This means that the semantic type of proper names must be lifted to the same level of complexity as that which is minimally required for quantifying phrases. Thus the proper name "John" must be represented as

$\lambda$ `R. R(john´)`

instead of simply `john´`, and the rule to compute the meaning of a sentence from the meaning of its constituent parts must be changed to become

`Sent* = Np*(Vp*)`

The same principle of lifting all phrases of a given syntactic type to the same semantic type must be applied throughout the entire grammar, which adds considerably to the complexity of the logical forms used as translations of phrases.

Second, since functional application assumes that the value of an argument is fully determined when a functor is applied to it we must often create lambda abstracts with several levels of embedding to get the desired behaviour. Thus, the transitive verb "beat" must be represented as

$\lambda$ `P.` $\lambda$ `Q. P(`$\lambda$ `Y. beat´(Y)(Q))`

Only now can we derive, compositionally, the meaning of the sentence "Peter beats John" from the meaning of its constituent parts: The translation of the transitive verb "beat" is functionally applied to the translation of the proper name "John"

```
λ P. λ Q. P(λ Y. beat´(Y)(Q))   (λ R. R(john´))
```

which thus represents the meaning of the verb phrase ''beats John'' (or ''being a John-beater''). This expression can be beta-reduced, in sequence, as follows:

```
λ Q. λ R. R(john´)(λ Y. beat´(Y)(Q))
λ Q. λ Y. beat´(Y)(Q)(john´)
λ Q. beat´(john´)(Q)
```

Functionally applying the meaning representation of 'Peter', which is, of course,

```
λ R. R(peter´)
```

to the meaning representation of ''beats John'' from above gives

```
λ R. R(peter´)( λ Q. beat´(john´)(Q))
```

which reduces, in sequence, as follows

```
λ Q. beat´(john´)(Q)(peter´),
beat´(john´)(peter´)
```

the latter of which is a notational variant of

```
beat´(peter´,john´),
```

which is the expected result.

If we have more complicated sentences (in particular, genuinely ambiguous ones, such as sentences with multiple quantifiers) this approach becomes nearly unmanageable. Moreover, it leads to many spurious syntactic ambiguities (''Peter smokes'' will, for instance, be ambiguous), which is about the last thing we need when we want to implement an NLP system. One way of getting almost the same effect but in a much simpler way is by sacrificing some formal elegance and by using *unification* instead of functional application to combine individual meaning representations. Then we can manipulate with impunity meaning representations which contain unbound variables (one of the charms of using the ''logical variables'' of Logic Programming languages). The use of this (by now common) technique allows us to use meaning representations of lexical items that are much simpler than Montagués while the rules combining constituent meanings get only marginally more complicated. By way of example, for mono-transitive verbs plus complement the rule translating syntax structures into logical forms is

```
sx2sm(cvp(_,Vb,PA,[],[],_,_),l(Ev,l(A,IVb & IPA)))
            :-    sx2sm(Vb,l(Ev,l(A,l(T2,IVb)))),
                  sx2sm(PA,l(T2,IPA)).
```

while the meaning of a transitive verb is now defined as

```
sx2sm(verb(Verb), l(Ev,l(T1,l(T2,@(Verb,Inst,Ev,T1,T2))))).
```

where `l(P,X)` stands for λ P.X and `@(F,...)` for F(...).

This method still allows a fully compositional semantic analysis. It also allows us to assert to the logical data base *partial logical forms* derived from subsentential phrases (after existential closure) in cases where we cannot parse a phrase in its entirety. In the given context of passage retrieval this means that in the limiting case, i.e. when *no* syntax structure can be recognised, the system behaves like a standard retrieval system: The individual *words* are turned into simple index or search terms. Such a system has thus, as a side-effect, the important property of ''graceful degradation'' or robustness: If a full linguistic analysis cannot be performed, the system falls back, step by step, to the behaviour of a standard information retrieval system (without statistical processing). Inevitably the precision of the system will degrade the more partial analyses occur but experience has shown that even in difficult cases at least noun phrases and prepositional phrases can be recognised, and the corresponding gain in precision over a traditional IR approach can be preserved even in those cases.

One last thing remains to be said about the semantic analysis part. Sometimes neither the pruning nor the filtering steps manage to narrow down the number of readings of a phrase to one (either because the phrase is genuinely ambiguous, or because our preference system cannot catch all the unintended readings). In such cases we must pass on all surviving readings to the semantic analysis module. There we create the *disjunction of all logical forms* and add the entire disjunction to the logical data base. This requires a slight extension of HCL as underlying logic, and the prover must also be modified accordingly.

*8.4 Performance*

LogDoc is a pure prototype system that was primarily used to test the *principles* of mixed-level representation and variable-depth evaluation. In particular, LogDoc was tested over no more than a few hundred titles of articles in the field of NLP. This means that its lexicon could be kept relatively small (slightly less than 60000 entries), its grammar very small (ca. 60 rules), and its semantic lexicon (meaning postulates, thesaurus, full term definitions) extremely small (a few dozen entries). Its parser was designed for ease of grammar development, not for speed, which resulted in parsing times of several minutes for phrases containing more than 10 word forms. Nevertheless, the main properties, viz. high precision, extensibility, robustness, and the use of variable-depth inference to increase recall without loss of precision, could be shown to be obtainable. However, it would clearly be meaningless to compare the performance of such a small system with that of other, much larger, systems. The question of whether LogDoc's principles could be applied profitably to larger amounts of text simply cannot be answered on the basis of the system as it stands now.

Work is now under way to extend LogDoc's coverage very considerably along different dimensions. The existing *grammar* is being replaced by a much larger grammar written in the dependency grammar tradition (viz. Link Grammar [22], [23]). A much faster *parser* for this grammar will be used than was available for LogDoc. For *semantic* resources we will be able to use a machine usable thesaurus, WordNet ([24]), as well as a sizeable machine usable lexicon with semantic information, Comlex ([25]). The amount of *text* available will also be much larger (all the, unedited, manual pages of the on-line Unix documentation). We hope to be able to show that the techniques described in this paper will continue to be useful in such an extended setting.

It is, however, clear that a direct and fair comparison between NLP based retrieval systems and standard information retrieval systems will probably not be possible for a long time to come. Standard IR systems are routinely used on document collections of hundreds of megabytes (and often gigabytes) of text, often from various sources and covering large domains. NLP systems capable of submitting similar amounts of text to a full-fledged linguistic analysis would require linguistic resources such as lexica, thesauruses, encyclopedias, and grammars with a coverage that is still far beyond anything available at present.

However, a direct comparison of NLP based retrieval systems with standard IR systems is not necessarily relevant as the two types of systems are not meant to be used for the same purpose. A typical use of an NLP based retrieval system, such as passage retrieval in a technical manual, will require the linguistic analysis of a few thousand, in extreme cases a few tens of thousands of pages of text. On the other hand, the precision required of such a system is far above anything required of an IR system. An IR system should therefore be seen rather as a front-end technology to NLP based retrieval systems, delivering pointers to candidate documents to be subjected to a closer linguistic analysis, than as a competing technology.

## 9. Related Work

The idea that a meaning representation language could be used as an indexing and retrieval language is not entirely new. There are a few systems that use a (very limited) amount of logic for retrieval (e.g. [26]). However, the notion of variable depth representation is less widespread. One system that is, on the surface, very similar to LogDoc in the way it uses logic to encode knowledge on different levels of generality, is SILOL by Sembok and Rijsbergen ([27]). In SILOL, too, documents are indexed by the logical representation of the noun phrases occurring in them. Queries are equally translated into logic. Documents are then retrieved in a way which is, at least functionally, equivalent to the kind of proof LogDoc performs. In order to cope with the fact that the semantic import of many syntactic relationships is either ambiguous or altogether unclear, Sembok and Rijsbergen introduce what they call ''generalised relationships''. As their name implies these relationships are intended to be so general as to cover the common components in such cases of ambiguity or vagueness. The common noun phrase *red horse* is therefore not translated as

```
horse(H)  ∧  red(H)
```

because this is correct only for the most common type of adjectival modification, the so-called *intersective* adjectives. However, this would not be the correct analysis for *intensionally* used adjectives, such as ''new'' in *new student*. This phrase can obviously not be translated as

```
student(S) ∧ new(S)
```

since we are not talking about things that are students and that are also new but about things that are new *as* students (''new'' modifies ''students''). For this reason, Sembok and Rijsbergen use the predicate `a(X,Y)` as common representation for *all* types of adjectival modification, which gives

```
horse(H) ∧ red(R) ∧ a(R,H)
```

Nominal compounding is translated in an analogous way: *system analysis* could mean either ''analysis of a system'' or ''analysis by a system''. It is therefore represented by the generalised relationship `r(X,Y)`, which gives the logical representation

```
system(S) ∧ analysis(A) ∧ r(S,A)
```

equivalent to our use of the predicate `by_with_for`. Surprisingly, the authors then go on to generalise this procedure and use it for *all* types of syntactic relationships. The relationship between subjects and verbs becomes `sv(X,Y)`, that between subject, verb and direct object `vso(X,Y,Z)` etc. Prepositions are treated as generalised relationships in their own right. However, this step reduces the power of their system quite considerably. The logical representation now encodes merely the syntactic structures themselves, not even part of their meaning. But without a logical model the whole problem of syntactic variability of natural language is back with a vengeance, and we could equally well save us the trouble of translating the syntax structures into the logical representation and use a direct syntax matching approach. It is, at least to us, not obvious why Sembok and Rijsbergen overstretched their approach to the point of it becoming nearly vacuous.

An approach that is closer to ours is that used in the system TACITUS ([19],[28]). TACITUS, too, represents unanalysable syntactic constructions by logical predicates. Thus a nominal compound such as ''lube-oil alarm'', which could mean any number of things is represented as

```
lubeoil(o) ∧ alarm(a) ∧ nn(o,a)
```

where `nn` denotes the semantically underspecified relationship between the component words (corresponding to `r` in [27] and to `by_with_for` in LogDoc). The authors then add axioms for the most common instantiations of this relationship, such as

```
∀ X,Y: part_of(X,Y)   → nn(X,Y)
∀ X,Y: sample(X,Y)    → nn(X,Y)
∀ X,Y: for(X,Y)       → nn(X,Y)
```

to cover examples like

**filter element**
**oil sample**
**lube-oil alarm**

Since the explanatory power of unanalysed relationships is lower than that of the more specific instantiations above, they are ascribed a low preference (given as a high cost factor in superscripts):

`lubeoil(o)` $^{\$5}$ ∧ `alarm(a)` $^{\$5}$ ∧ `nn(o,a)` $^{\$20}$

TACITUS then uses *weighted abduction* to find the cheapest consistent set of assumptions needed to explain a given utterance. To control the potentially explosive nature of abduction the authors use a type hierarchy that excludes semantically aberrant assumptions.

As far as can be ascertained from the publicly available information on TACITUS, weighted abduction amounts to a kind of variable-depth evaluation similar to ours. The main differences between TACITUS and LogDoc are that LogDoc uses straightforward deduction rather than abduction, and that LogDoc makes a more systematic distinction between the levels of specificity in the knowledge representation. It remains unclear to what extent TACITUS uses other types of information to control the inference process (such as the length of proofs used in LogDoc).

It seems that TACITUS was superseded by work on a much simpler, linguistically less interesting, but much faster system, FASTUS ([29]). The reason why we think it worthwhile to continue work along the lines described above is that we want to get incrementally closer to real *question answering over texts*, and for this we will eventually need full linguistic analysis of texts, something that is not required for the extraction and routing tasks tested in the Message Understanding Conferences (MUC) for which FASTUS was developed.

**Acknowledgements**

This work has been funded by the Swiss National Science Foundation (projects 4023-026996 and 1214-045448.95/1)